\documentclass[sigconf]{acmart}

\usepackage[utf8]{inputenc} 
\usepackage[T1]{fontenc}    
\usepackage{hyperref}       
\usepackage{url}            
\usepackage{booktabs}       
\usepackage{amsfonts}       
\usepackage{nicefrac}       
\usepackage{microtype}      
\usepackage{xcolor}         

\usepackage{multirow}
\usepackage{subfigure}

\usepackage[english]{babel}
\usepackage{moresize}
\usepackage{amsmath}
\usepackage{algorithmic}
\usepackage{balance}
\usepackage{comment}
\usepackage{paralist}
\usepackage{bm}
\usepackage{pgfplots}
\usetikzlibrary{pgfplots.dateplot}

\usepackage{flushend}
\usepackage[english]{babel}
\usepackage{graphicx}

\usepackage{amssymb}
\usepackage{amsfonts}
\usepackage{url}
\usepackage{bbm}
\usepackage{longtable}
\usepackage{rotating}
\usepackage{multirow}
\usepackage{mathrsfs}
\usepackage{enumitem}
\usepackage[linesnumbered,algoruled,boxed,lined]{algorithm2e}
\usepackage{adjustbox}
\usepackage{hyperref}
\usepackage{pgfplots}
\usetikzlibrary{pgfplots.dateplot}
\usepackage{filecontents}

\newcommand{\eg}{\textit{e}.\textit{g}.} 
\newcommand{\ie}{\textit{i}.\textit{e}.}

\newcommand{\mata}{\textbf{A}}
\newcommand{\matd}{\textbf{D}}
\newcommand{\mate}{\textbf{E}}
\newcommand{\mati}{\textbf{I}}
\newcommand{\matn}{\textbf{N}}

\newcommand{\matw}{\textbf{W}}

\newcommand{\vece}{\textbf{e}}
\newcommand{\vech}{\textbf{h}}

\newcommand{\vecx}{\textbf{x}}
\newcommand{\vecb}{\textbf{b}}

\newcommand{\setu}{\mathcal{U}}
\newcommand{\setv}{\mathcal{V}}
\newcommand{\sete}{\mathcal{E}}
\newcommand{\seth}{\mathcal{H}}
\newcommand{\graph}{\mathcal{G}}

\newcommand{\param}{\mathbf{\Theta}}
\newcommand{\loss}{\mathcal{L}}
\newcommand{\gauss}{\mathcal{N}}
\newcommand{\calo}{\mathcal{O}}
\def\model{DiffGraph}

\AtBeginDocument{%
  }

\copyrightyear{2025}
\acmYear{2025}
\setcopyright{acmlicensed}\acmConference[WSDM '25]{Proceedings of the Eighteenth ACM International Conference on Web Search and Data Mining}{March 10--14, 2025}{Hannover, Germany}
\acmBooktitle{Proceedings of the Eighteenth ACM International Conference on Web Search and Data Mining (WSDM '25), March 10--14, 2025, Hannover, Germany}
\acmDOI{10.1145/3701551.3703590}
\acmISBN{979-8-4007-1329-3/25/03}

\begin{document}
\title{DiffGraph: Heterogeneous Graph Diffusion Model}

\author{Zongwei Li}
\affiliation{%
  \institution{University of Hong Kong}
  \city{Hong Kong}
  \country{China}
}
\email{conwayli@tencent.com}
\orcid{0009-0000-3112-1268}

\author{Lianghao Xia}
\affiliation{%
  \institution{University of Hong Kong}
  \city{Hong Kong}
  \country{China}
  }
\email{aka\_xia@foxmail.com}
\orcid{0000-0003-0725-2211}

\author{Hua Hua}
\affiliation{%
  \institution{Tencent}
  \city{Shenzhen}
  \state{Guangdong}
  \country{China}
}
\email{mikehua@tencent.com}
\orcid{0000-0003-4942-5395}

\author{Shijie Zhang}
\affiliation{%
  \institution{Guangdong Engineering Center for Social Computing and Mental Health}
  \city{Shenzhen}
  \state{Guangdong}
  \country{China}
}
\email{zhang.shijie1101@gmail.com}
\orcid{0000-0003-3226-6842}

\author{Shuangyang Wang}
\affiliation{%
  \institution{Tencent}
  \city{Shenzhen}
  \state{Guangdong}
  \country{China}
}
\email{feymanwang@tencent.com}
\orcid{0000-0002-6180-8607}

\author{Chao Huang}
\authornote{Chao Huang is the Corresponding Author.}
\affiliation{%
  \institution{University of Hong Kong}
  \city{Hong Kong}
  \country{China}
  }
\email{chaohuang75@gmail.com}

\renewcommand{\shortauthors}{Zongwei Li, Lianghao Xia, Hua Hua, Shijie Zhang, Shuangyang Wang, \& Chao Huang}

\begin{abstract}
Recent advances in Graph Neural Networks (GNNs) have revolutionized graph-structured data modeling, yet traditional GNNs struggle with complex heterogeneous structures prevalent in real-world scenarios. Despite progress in handling heterogeneous interactions, two fundamental challenges persist: noisy data significantly compromising embedding quality and learning performance, and existing methods' inability to capture intricate semantic transitions among heterogeneous relations, which impacts downstream predictions. To address these fundamental issues, we present the Heterogeneous Graph Diffusion Model (\model), a pioneering framework that introduces an innovative cross-view denoising strategy. This advanced approach transforms auxiliary heterogeneous data into target semantic spaces, enabling precise distillation of task-relevant information. At its core, \model\ features a sophisticated latent heterogeneous graph diffusion mechanism, implementing a novel forward and backward diffusion process for superior noise management. This methodology achieves simultaneous heterogeneous graph denoising and cross-type transition, while significantly simplifying graph generation through its latent-space diffusion capabilities. Through rigorous experimental validation on both public and industrial datasets, we demonstrate that \model\ consistently surpasses existing methods in link prediction and node classification tasks, establishing new benchmarks for robustness and efficiency in heterogeneous graph processing. The model implementation is publicly available at: \textcolor{blue}{\url{https://github.com/HKUDS/DiffGraph}}.
\end{abstract}



\maketitle

\section{Introduction}
\label{sec:intro}

Learning with heterogeneous graphs has emerged as a pivotal paradigm in modern data science, reflecting the complexity of real-world systems. Unlike homogeneous graphs with uniform node and edge types, heterogeneous graphs encapsulate rich, multi-faceted interactions among diverse entities, enabling more expressive data representations. These advanced graph structures have demonstrated remarkable effectiveness across numerous domains, from bibliographic academic data~\cite{hu2019strategies, wang2019heterogeneous}, medical data~\cite{davis2017comparative}, to recommender systems~\cite{yang2022multi}. The fundamental objective lies in leveraging this inherent diversity to enhance various analytical tasks. By harnessing these heterogeneous relationships, researchers have achieved significant breakthroughs in node classification~\cite{tian2023heterogeneous}, link prediction~\cite{fan2019metapath}, and graph classification~\cite{sun2021heterogeneous}, consistently demonstrating superior performance compared to traditional homogeneous approaches.

Recent years have witnessed remarkable advancements in heterogeneous graph neural networks (HGNNs). Through relation-aware message passing frameworks, earlier study~\cite{zhang2019heterogeneous} has improved graph learning tasks by effectively capturing both complex structural patterns and diverse semantic information within heterogeneous graphs. Notable research works include HGT~\cite{tian2023heterogeneous}, which leverages graph attention mechanisms to dynamically assess the importance of various heterogeneous paths, thereby enhancing both semantic-level and node-level representation learning. Similarly, HeteGNN~\cite{zhang2019heterogeneous} introduces a framework that simultaneously models heterogeneous structures and their associated content features. Building upon the success of self-supervised learning (SSL)~\cite{misra2020self, you2020graph} in addressing data scarcity and noise challenges, recent research efforts\cite{wang2021self, tian2023heterogeneous,li2024recdiff} have effectively incorporated SSL techniques to further advance heterogeneous graph learning capabilities.

Despite significant advances in heterogeneous graph learning, two critical challenges remain insufficiently addressed. \textbf{First}, current HGNN approaches demonstrate limited capability in handling heterogeneous noise. While self-supervised learning techniques, particularly contrastive learning~\cite{wang2021self}, attempt to mitigate data noise through self-supervisory signals, they lack sophisticated mechanisms for processing noisy data with complex heterogeneous semantics. These methods primarily rely on simple random data augmentations (\eg, feature masking, random walk), overlooking the crucial interactions between different relation types that could significantly impact noise distribution. \textbf{Second}, existing approaches struggle to effectively model complex transition patterns among heterogeneous relations. Although some methods employ attention mechanisms~\cite{xia2020multiplex} or meta paths~\cite{fu2020magnn} to capture dependencies, these approaches are insufficient for encoding sophisticated semantic transition processes between heterogeneous relation types, potentially compromising the extraction of task-relevant information and degrading downstream prediction performance. These observations lead us to two fundamental research questions:
\begin{itemize}[leftmargin=*]
    \item How can we develop a robust learning paradigm that effectively addresses and mitigates data noise in heterogeneous graphs?\\\vspace{-0.12in}
    \item How can we accurately capture the complex semantic transition processes across diverse relation types in heterogeneous graphs?
\end{itemize}
To address these challenges, we propose a dual-component solution. \textbf{First}, we introduce an adaptive parametric function that systematically filters noisy structures while preserving task-critical information for downstream predictions. This function is specifically designed to encode heterogeneous semantics and structural patterns, enabling effective characterization of complex noise distributions. \textbf{Second}, we develop an advanced semantic transition model to capture intricate relationships across different relation types, employing a fine-grained approach to accommodate diverse heterogeneous semantic patterns and their evolutionary processes.

Inspired by the remarkable capabilities of diffusion models in capturing complex data generation processes, we propose a novel generative diffusion framework for heterogeneous graphs. Our approach implements a bi-directional process: a forward phase for systematic noise addition and a backward phase for noise removal, jointly enhancing model robustness against heterogeneous data noise. Extending beyond conventional diffusion models, our method uniquely facilitates the transfer between auxiliary and target graph views through multiple fine-grained diffusion steps, enabling precise modeling of semantic transitions at a granular level.

Following these ideas, we present a Heterogeneous Graph Diffusion Model (\model). Our approach begins by identifying the target subgraph containing node and edge types most crucial to the prediction task, while designating the remaining structure as the auxiliary graph for feature enhancement. A specialized heterogeneous graph encoder, built on graph convolutions, projects both target and auxiliary graphs into a latent representation space. We then introduce a novel latent heterogeneous graph diffusion module that orchestrates forward and backward diffusion processes on these encoded representations, with auxiliary graph data serving as the diffusion source and target graph data providing denoising training signals. Our proposed diffusion framework accomplishes two key objectives: capturing complex noise distributions in heterogeneous graph data and modeling semantic transitions between diverse relation types. By conducting noise addition and removal operations in latent space rather than the original graph space, \model effectively addresses the challenges of generating sparse and discrete heterogeneous graph data. This design choice significantly enhances the model's capability for unbiased heterogeneous graph modeling, overcoming traditional limitations in graph generation. The main contributions of this paper can be summarized as follows:
\begin{itemize}[leftmargin=*]
    \item We propose the \model, a novel approach that enhances model performance by systematically filtering non-essential semantic information from heterogeneous graph structures.\\\vspace{-0.12in}
    \item We develop an innovative hidden-space diffusion mechanism that effectively removes noisy information through a sophisticated multi-step process of controlled noise addition and denoising within heterogeneous embeddings.\\\vspace{-0.12in}
    \item We present extensive experimental evaluations of \model across diverse datasets, demonstrating its superior effectiveness in both link prediction and node classification tasks.
\end{itemize}

\section{Methodology}
\label{sec:method}
\begin{figure*}[t]
    \centering
    
    \includegraphics[width=\textwidth]{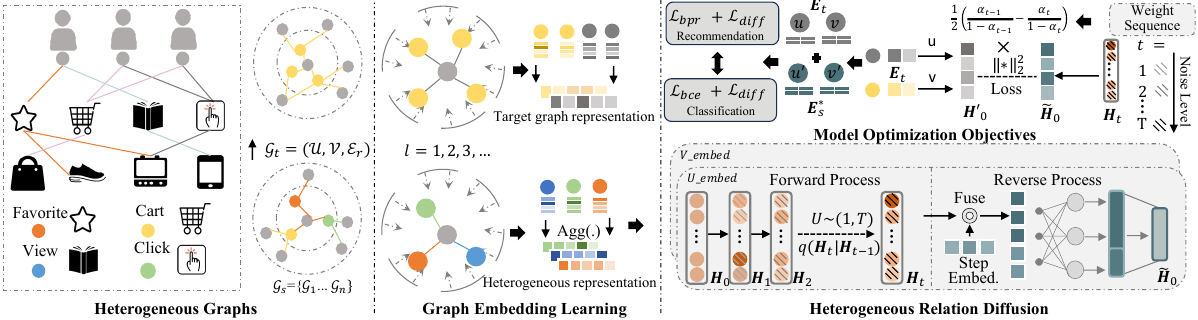}
    \vspace{-0.2in}
    \caption{Overall architecture of the proposed \model\ framework.}
    \label{fig:framework}
    \vspace{-0.15in}
\end{figure*}

This section outlines the details of the proposed \model\ framework. The architecture of \model\ is illustrated in Figure~\ref{fig:framework}.

\subsection{Heterogeneous Graph Learning}
\noindent\textbf{Heterogeneous Graphs}. In a heterogeneous graph, the node set is denoted as $\mathcal{V} = \{v_i\}$, with the set of node types denoted as $\mathbb{V}$. There exists a node type mapping function $\phi: \mathcal{V} \to \mathbb{V}$, where $\phi(v_i)$ represents the type of node $v_i$. Similarly, the set of edges is denoted as $\mathcal{E} = \{(v_i, v_j)\}$, with an edge type mapping function $\psi: \mathcal{E} \to \mathbb{E}$. Here, $\mathbb{E}$ denotes the set of edge types, and $\psi(v_i,v_j)$ represents the type of edge $(v_i, v_j)$. With these notations, we formally define the heterogeneous graph as $\mathcal{G} = (\mathcal{V}, \mathcal{E}, \phi, \psi)$. For convenience, a heterogeneous graph is recorded by the binary adjacency matrix $\textbf{A}$ of size ${|\mathcal{V}| \times |\mathcal{V}| \times |\mathbb{E}|}$. An entry $a_{i,j}^r \in \textbf{A}$ equals 1 if there is a link of type $r \in \mathbb{E}$ between nodes $v_i$ and $v_j$, otherwise $a_{i,j}^r = 0$.\\\vspace{-0.12in}

\noindent \textbf{Heterogeneous Graph Prediction}.
A predictive model $f$ for heterogeneous graphs can be divided into an encoding phase and a prediction phase, formally as: $f(\mathcal{G}) = \text{Pred} \circ \text{Enc}(\mathcal{G})$. The encoding phase $\text{Enc}(\cdot)$ learns $d$-dimensional hidden representations $\textbf{E} \in \mathbb{R}^{|\mathcal{V}| \times d}$ for nodes $v_i \in \mathcal{V}$, while the prediction phase $\text{Pred}(\cdot)$ generates task-specific predictions based on the learned embeddings.
Typical graph forecasting tasks include node classification and link prediction. For node classification, the prediction network $\text{Pred}(\cdot)$ of \model\ employs a multilayer perceptron (MLP) to infer nodes' classes from their learned embeddings. For link prediction, the dot-product operator is utilized to infer the existence of edges based on the embeddings of the connected nodes in $\text{Pred}(\cdot)$.

Motivated by the simplicity and effectiveness of Graph Convolutional Networks (GCNs)~\cite{gao2018large}, our \model\ adopts a GCN-based encoding process for $\text{Enc}(\cdot)$, which is formally defined as follows:
\begin{align}
    \label{eq:rec_enc}
    &\mate= \text{Pooling}(\{\mate_r|r\in\mathbb{E}\}),~~~~ \mate_r = \sum\nolimits_{l=0}^L \mate_{r,l}\nonumber\\
    &\mate_{r,l} = \text{norm}(\delta(\matd^{-1/2}\mata^r\matd^{-1/2}\mate_{r,{l-1}}))
\end{align}
where $\mathbf{A}^r \in \mathbb{R}^{|\mathcal{V}| \times |\mathcal{V}|}$ denotes the specific adjacency matrix for each heterogeneous relation $r \in \mathbb{E}$, and $\mathbf{D}$ denotes its corresponding diagonal degree matrix. $\mathbf{E}_{r,l} \in \mathbb{R}^{|\mathcal{V}| \times d}$ represents the embedding matrix at the $l$-th GCN iteration. For datasets without node features, the initial embeddings $\mathbf{E}_{r,0}$ are randomly initialized learnable parameters. For datasets with node features, $\mathbf{E}_{r,0}$ utilizes the initial node features. During each iteration, the existing embeddings are disseminated to adjacent nodes through the Laplacian-normalized adjacency matrix, utilizing the activation function $\delta(\cdot)$ and the $l_2$ embedding normalization function $\text{norm}(\cdot)$. Following $L$ iterations, the final node embeddings $\mathbf{E}_r$ are generated from an element-wise aggregation of multi-order node embeddings. And the final node embeddings $\mate$ is obtained by aggregate node embeddings $\mate_r$ of different relation types with a pooling function.

\subsection{Cross-view Heterogeneous Graph Denoising}
\subsubsection{\bf Denoising Auxiliary Graphs using Target Graph}
To enhance the performance of heterogeneous graph prediction, we propose a cross-view denoising approach to maximally extract task-relevant information from heterogeneous graphs. Initially, we identify the task-relevant subgraph within the compound heterogeneous graph as the target graph $\mathcal{G}_t$. An example of such a target graph is the purchase behavior graph between users and items in an e-commerce recommender system. Once this target graph $\mathcal{G}_t$ is removed from the entire graph $\mathcal{G}$, the remaining graph is referred to as the auxiliary graph $\mathcal{G}_s$ (also known as the source graph).

Our cross-view denoising framework for heterogeneous graphs aims to maximize the mutual information between the auxiliary view and the target view by learning a parametric function $g$ that maps the auxiliary graph space to the task-relevant data space. This learnable denoising function $g$ takes the source graph data $\mathcal{G}_s$ as the input and utilizes the target graph data $\mathcal{G}_t$ as labels. Specifically, this denoising framework can be formulated as follows:
\begin{align}
    \mathop{\arg\min}\nolimits_{\param_f, \param_g}~~\loss_{main}(\mathcal{Y}, f(\mathcal{G})) + \lambda\cdot\loss_{deno}(\graph_t, g(\graph_s))
\end{align}
where $\mathcal{L}_{main}$ represents the loss for the main prediction task, such as BPR loss~\cite{rendle2009bpr} for link prediction and cross-entropy loss for node classification. Here, $\mathcal{Y}$ denotes the task-specific labels. Meanwhile, $\mathcal{L}_{deno}$ denotes the learning objective for our denoising function $g$, which measures the differences between the task-relevant target graph $\mathcal{G}_t$ and the denoising output of function $g$. The learning process tunes the parameter sets $\theta_f$ and $\theta_g$, which are parameters for the forecasting model $f$ and the denoising network $g$, respectively. A weighting parameter $\lambda$ is used to balance the denoising task.

\subsubsection{\bf Latent Heterogeneous Graph Diffusion}
Inspired by the impressive performance of diffusion models in capturing complex data generation processes across diverse data types~\cite{Croitoru_Hondru_Ionescu_Shah_2022}, we propose to instantiate our denoising function as a heterogeneous graph diffusion model. This model aims to distill task-relevant information from the auxiliary graph by iteratively refining the graph data. Given the complexity of graph-structured data, we propose conducting this diffusion process in the latent representation space of the heterogeneous graph data. Specifically, our diffusion model aims to achieve the following step-by-step transformation:
\begin{align}
    \label{eq:hidden_project}
    \graph_s^{*} \stackrel{\pi}{\rightleftharpoons} \mate_s^{*} \stackrel{\varphi}{\longrightarrow} \tilde{\mate}_s^{*} \stackrel{\varphi'}{\longrightarrow} \hat{\mate}_s^{*} \stackrel{\pi'}{\rightleftharpoons} \hat{\graph}_s^{*},~~~~\graph_s^{*}=\{\graph_s^{1}, ..., \graph_s^{n}\}
\end{align}
where $\pi$ and $\pi'$ denote the bidirectional mapping between the graph-structured data and the latent representations. By utilizing the heterogeneous GCN-based encoder in the prediction function $f$, $\mathbf{E}_s^{*}$ captures heterogeneous semantic information from the source view. Function $\varphi$ represents the forward diffusion process, which incrementally adds noise to $\mathbf{E}_s^{*}$. Correspondingly, $\varphi'$ refers to the reverse diffusion process, which removes the noise step by step.\\\vspace{-0.12in}

\noindent \textbf{Rationale for Latent Diffusion over Graph Diffusion}. In the context of graph-based task learning, the role of the graph encoder is to aggregate the embeddings of neighboring nodes, transforming the structural semantics of the graph into low-dimensional information within the latent space. This characteristic suggests that by learning the denoising process $\varphi'$ within the latent space, our hidden-space diffusion model can effectively filter out the imprecise semantic differences within heterogeneous graphs, thereby better supporting the learning of the target graph.

Moreover, graph data exhibits several challenging characteristics that hinder effective diffusion, including its sparse and discrete nature, as well as its large data space of exponential complexity. In contrast, the latent space is dense and continuous, and as a compressed representation of the graph, it usually has a smaller size. These key advantages motivate us to conduct graph diffusion in the latent representation space, instead of in the graph space.

\subsubsection{\bf Forward and Reverse Diffusion}
\noindent\textbf{Forward Process}. The forward process of our diffusion model consists of $T$ steps of gradual noise addition. This process begins with the encoded hidden representations, denoted as $\mathbf{H}_0 = \mathbf{E}_s^{*}$. It then iteratively increases the noise ratio following the recursive formula below:
\begin{align}
    \label{eq:gauss_step}
    q(\seth_t|\seth_{t-1}) = \gauss(\seth_t; \sqrt{1-\beta_t} \seth_{t-1}, \beta_t \mati)
\end{align}
where $\beta_t$ controls the noise scale. $\mathcal{N}$ denotes the Gaussian distribution used to generate the noise, and the noise data $\mathbf{H}_t$ will progressively increase as $\beta_t$ increases, ultimately converging to pure Gaussian noise~\cite{ho2020denoising}. This property allows our noise resolution methodology to cover a wide range of noise intensities within the training data. The additive nature of the Gaussian distribution enables us to directly compute the data at step $t$ using $\mathbf{H}_0$ and pre-computed values related to the $\beta_t$ sequence. Specifically, we can pre-calculate the following values to expedite the diffusion process:
\begin{align}
    \label{eq:alpha_gen}
     \alpha_t=1-\beta_t,~~~~\bar{\alpha}_t=\prod_{t'=1}^T\alpha_{t'},~~~~\beta_t = 1 - b_t / b_{t-1}
\end{align}
To determine $\beta_t$ at each step, we introduce two hyperparameters, $\bar{b}_{max}$ and $\bar{b}_{min}$. These parameters generate a linear interpolation sequence $b = (1, \bar{b}_{max}, \cdots, \bar{b}_{min})$. Using the calculations from Eq~\ref{eq:alpha_gen}, we derive the following results:
\begin{align}
    \seth_t &= \sqrt{\alpha_t} \seth_{t-1} + \sqrt{1-\alpha_t} \xi_1 \nonumber\\
    &\Rightarrow\sqrt{\alpha_t}(\sqrt{\alpha_{t-1}} \seth_{t-2} + \sqrt{1-\alpha_{t-1}} \xi_2 ) + \sqrt{1-\alpha_t} \xi_1\nonumber\\
    &\Rightarrow\sqrt{\bar{\alpha}_t} \seth_0 + \sqrt{1-\bar{\alpha}_t}{\xi}'_t,~~~~\xi\mapsto\mathcal{N}(0, \mati)
\end{align}
where $\xi_{*}$ represents independent Gaussian distributions. By pre-calculating ${\alpha_t}$, we can leverage the additive properties to infer $\mathcal{H}_t$ efficiently, bypassing the need for recursive computations.\\\vspace{-0.12in}

\noindent\textbf{Reverse Process}.
The purpose of the reverse process is to enable \model\ to learn the denoising ability of noisy heterogeneous semantic data. Specifically, we aim to reconstruct the heterogeneous relations in the hidden space from the noisy data $\seth_t$. A learnable neural network is employed to estimate the generation of conditional probabilities for the following:
\begin{align}
    \label{eq:gaus_output}
    p(\seth_{t-1}|\seth_t)=\gauss\left(\seth_{t-1}; \mathbf{\mu}_\theta(\seth_t, t),\mathbf{\Sigma}_\theta(\seth_t,t)\right)
\end{align}
where $\mathbf{\mu}_\theta(\cdot)$ and $\mathbf{\Sigma}_\theta(\cdot)$ are obtained from a two-layer neural network with learning parameter $\theta$, which is intended to be used for estimating Gaussian distributions. The framework is defined:
\begin{align}
    \label{eq:gaus_network}
    \mathbf{\mu}_\theta(\vech_t,t)=\textbf{MLP}^2(\vech_t\| \textbf{s}_t),~~\textbf{MLP}(\textbf{x})=\sigma(\matw\vecx+\vecb)
\end{align}
where $\textbf{MLP}^2(\cdot)$ represents two continuous MLP layers. We concatenate the t-step hidden vector $\vech_t$ with a time step-specific embedding $\textbf{s}_t$ input data $x$. $\sigma$, $\matw$ and $\vecb$ denotes the activation function, linear transformation and bias for the neural network respectively.

\subsection{Model Training for \model}
\subsubsection{\bf Diffusion Loss Function}
To optimize the hidden-space diffusion module,  we adopt the embedding 
$\mate_{t}$ of the target graph as the evidence lower bound (ELBO). Based on the data specific to each side, $\vech_{0}^{'}=\mate_u,\mate_v \in \mate_t$, the paradigm for maximizing the ELBO can be summarized as follows:
\begin{align}
    &\log p(\vech_{0}^{'})\geq \underbrace{\mathbb{E}_{q\left(\vech_{1} | \vech_{0}\right)}\left[\log p_{\theta}\left(\vech_{0}^{'} | \vech_{1}\right)\right]}_{(\text {reconstruction term) }}\\
    &-\sum_{t=2}^{T} \underbrace{\mathbb{E}_{q\left(\vech_{t} | \vech_{0}\right)}\left[D_{\mathrm{KL}}\left(q\left(\vech_{t-1} | \vech_{t}, \vech_{0}\right) \| p_{\theta}\left(\vech_{t-1} | \vech_{t}\right)\right)\right]}_{\text {(denoising comparing term) }}\nonumber
\end{align}
The model incorporates two optimization terms. The objective of the denoising comparising term is to align the true distribution $q\left(\vech_{t-1} | \vech_{t}, \vech_{0}\right)$ more approximate our heterogeneous semantics denoiser $ p_{\theta}\left(\vech_{t-1} | \vech_{t}\right))$, thereby minimizing the KL divergence. Drawing on the conclusions of previous works~\cite{wang2023diffusion,ho2020denoising}, we can simplify the study of the standard deviation as $\mathbf{\Sigma}_\theta(\vech_t,t)=\sigma^2(t)\mati$. The $\loss_t$ for denoising comparising term can be transformed as follow:
\begin{align}
    \loss_t=\mathbb{E}_{q(\vech_t|\vech_0)}\left[\frac{1}{2}\left(\frac{\bar{\alpha}_{t-1}}{1-\bar{\alpha}_{t-1}} - \frac{\bar{\alpha}_t}{1-\bar{\alpha}_t} \right) \|\hat{\vech}_\theta(\vech_t,t)-\vech_0^{'}\|_2^2 \right]
\end{align}
where $\hat{\vech}_\theta(\vech_t,t)$ denotes the data $\vech_0$ predicted by $\vech_t$ and $t$ via a Multi-Layer Perceptron (MLP), which indicated in Eq~\ref{eq:gaus_network}.
From the inference of Eq\ref{eq:gaus_network}, the reconstruction term can also be approximated as the squared loss between the optimization $\hat{\vech}_\theta(\vech_t,t)$ and the target graph vector $\vech_0^{'}$ . We define it as:
\begin{align}
    \label{eq:squared_loss}
    \mathcal{L}_1=\mathbb{E}_{q(\vech_t|\vech_0)}\left[ \|\hat{\vece}_\theta(\vech_1,1)-\vech_0^{'}\|_2^2 \right]
\end{align}
It is noteworthy that the semantic information in the latent space of the source heterogeneous graphs serves as the input for the diffusion module, and the denoised embedding is optimized by forming a control group with the semantic vector $\vech_0^{'}$ of the target graph. In determining the number of steps $t$, we employ a uniform sampling strategy. Formally, the diffusion loss $\mathcal{L}_{deno}$ is as follows:
\begin{align}
    \label{eq:diff_loss}
    \mathcal{L}_{deno}=\mathbb{E}_{t \sim \matn (1,t)}\mathcal{L}_{t}
\end{align}

\subsubsection{\bf Prediction and Optimization}
For the link prediction task, we combine the denoised heterogeneous semantic relations of the source graphs with the semantic relations of the target graph to get the final embedding to make predictions. And we employ the BPR loss function to optimize predictions $\hat{{y}}$:
\begin{align}
    \label{eq:fused_emb}
    \hat{{y}}_{i,j} &= \tilde{\vece}_i^\top \tilde{\vece}_j,~~~~~~\tilde{\mate} = \mate^t + \hat{\mate}^{*},\nonumber\\
    \loss_{main}&=\sum_{u,v^+,v^-} -\log \text{sigm}(\hat{y}_{u,v^+} - \hat{y}_{u,v^-})
\end{align}
where ${u,v^+,v^-}$ stands for the positive and negative sample sampling strategy~\cite{rendle2009bpr}. For the node classification task, we utilize the following cross entropy loss function:
\begin{align}
    p_{i,c}=\text{softmax}(\text{MLP}(\tilde{\vece}_i))_c,~~~~  \loss_{main} = - \sum_{v_i\in\mathcal{V}} \log p_{i,y_i}
\end{align}

\subsubsection{\bf Model Complexity Analysis}
This section gives a thorough analysis on the time and space complexity of our \model.

\noindent\textbf{Time Complexity}. Initially, \model\ performs graph-level information propagation on both the target collaborative graph $\graph_t$ and the auxiliary heterogeneous graphs $\graph_{*}=\{ \graph_1, ...,  \graph_n, ..., \graph_N \}$. This standard graph convolutional process requires $\calo((|\sete_r|+\sum_{l=1}^N |\sete_n|)\times d)$ calculations for message passing and $\calo((|\setu|+|\setv|)\times d^2)$ for embedding transformation. For recommedation task, our diffusion process operates on the user side and item side at each training epoch respectively. The forward diffusion process costs $\calo(|\setu|\times d)$ and $\calo(|\setv|\times d)$ computations, while the reverse process costs $\calo(|\setu|\times (d^2 +dd'))$ and $\calo(|\setv|\times (d^2 +dd'))$.For node classification task, the diffusion process only performs on user-tie nodes, which costs  $\calo(|\setu|\times d)$ for forward diffusion process and $\calo(|\setu|\times (d^2 +dd'))$ for reverse process.  
In conclusion, \model\ achieves comparable time costs to common heterogeneous recommenders.

\noindent\textbf{Memory Complexity}. The graph encoding process of our \model\ model requires a similar number of parameters as conventional graph-based heterogeneous models. The hidden-space diffusion network employs $\calo(d^2+dd')$ parameters for the denoiser. 
\section{Evaluation}
\label{sec:eval}
We evaluate the performance of our \model\ framework by studying the following Research Questions (RQs):
\begin{itemize}[leftmargin=*]
    \item \textbf{RQ1}: How does the proposed \model\ framework perform on link prediction, node classification, and industry datasets?
    \item \textbf{RQ2}: How effective are the designed modules in \model?
    \item \textbf{RQ3}: How do different settings of key hyperparameters impact the graph prediction accuracy of our \model\ method?
    \item \textbf{RQ4}: How is the efficiency of \model\ compared to baselines?
    \item \textbf{RQ5}: How effectively can our \model\ approach perform to alleviate the issue of data sparsity for graph data?
    \item \textbf{RQ6}: How well can \model\ handle noisy heterogeneous graphs?
    \item \textbf{RQ7}: Can \model\ provide explanations in specific cases?
\end{itemize}

\begin{table}[t]
    \centering
    \caption{Statistics of experimental datasets.}
    \label{tab:datasets}
    \vspace{-0.12in}
    \small
    \setlength{\tabcolsep}{0.6mm}
        \begin{tabular}{c|c|c|c|c}
            \hline
            Dataset       & User \# & Item \# & Link \# & Interaction Types                       \\ \hline
            Tmall         & 31882   & 31232   & 1,451,29      & View, Favorite, Cart, Purchase \\ \hline
            Retail Rocket & 2174    & 30113   & 97,381        & View, Cart, Transaction        \\ \hline
            IJCAI         & 17435   & 35920   & 799,368       & View, Favorite, Cart, Purchase \\ \hline
            Industry        & 1M   & 361   & 23,890,445       & Purchase, Friend, Complete Task \\ \hline
            
        \end{tabular}%
    {
    \vspace{0.15in}
    \begin{tabular}{c|c|c||c|c|c}
    & Node & Metapath & \multirow{5}{*}{AMiner} & Node & Metapath\\ \hline
    \multirow{4}{*}{DBLP} & \textbf{A}uthor:4057 & APA & &\textbf{P}aper:6564 & PAP\\
     & \textbf{P}aper:14328 & APCPA & &\textbf{A}uthor:13329 & PRP\\
     & \textbf{C}onference:20 & APTPA & &\textbf{R}eference:35890 & POS\\
     & \textbf{T}erm:7723 &  &&\\
     \hline
    \end{tabular}
    }
    \vspace{-0.2in}
\end{table}

\subsection{Experimental Settings}
\subsubsection{\bf Datasets} We evaluate \model\ on link prediction and node classification tasks. For link prediction, we utilize \textbf{Tmall}, \textbf{Retailrocket}, and \textbf{IJCAI} datasets. For node classification, we employ \textbf{DBLP} and \textbf{AMiner} (focusing on academic networks), along with an \textbf{Industry} dataset from a gaming platform. Detailed statistics are presented in Table~\ref{tab:datasets}, with dataset descriptions below.\\\vspace{-0.12in}

\noindent\textbf{Link Prediction Dataset}: $\bullet$ \textbf{Tmall}: An E-commerce dataset with user views, favorites, cart additions, and purchases, filtered to users with $\geq$3 purchases~\citep{xia2020multiplex}. $\bullet$ \textbf{Retailrocket}: A dataset with user page views, cart additions, and transactions, where purchases are primary and other interactions auxiliary~\cite{yang2022multi}. $\bullet$ \textbf{IJCAI}: A dataset from IJCAI15 with heterogeneous user behaviors similar to Tmall.

\noindent\textbf{Node Classification Dataset}: $\bullet$ \textbf{Industry}: A game platform dataset with retention as target behavior and three auxiliary behaviors (purchases, friendships, tasks), forming user-item, user-user, and user-task graphs. $\bullet$ \textbf{DBLP}: A dataset subset of authors categorized into four research areas (Database, Data Mining, AI, and Information Retrieval) for node classification. $\bullet$ \textbf{AMiner}: A dataset containing papers in four categories, utilizing meta-path relationships (author, paper, conference, reference) for heterogeneous node classification.

\subsubsection{\bf Baseline Methods} 
For a thorough evaluation of \model, we include the following 21 baselines from different research lines.\\\vspace{-0.12in}

\noindent\textbf{Recommendation Models with Heterogeneous Relations}:\vspace{-0.05in}
\begin{itemize}[leftmargin=*]
    \item \textbf{NMTR} \cite{gao2019neural}: It uses a multi-task learning framework with redefined cascading relationships to explore diverse user behaviors.
    \\\vspace{-0.12in}
    \item \textbf{MATN} \cite{xia2020multiplex}: This method uses a memory-augmented attention mechanism for the modeling of multiple user behaviors. 
    \\\vspace{-0.12in}
    \item \textbf{MBGCN} \cite{jin2020multi}:  It utilizes a GCN-based model to analyze multiple patterns of behavior within the user-item interaction graph.
\end{itemize}

\noindent\textbf{Heterogeneous Graph Neural Networks}:\vspace{-0.05in}
\begin{itemize}[leftmargin=*]
    \item \textbf{Mp2vec} \cite{dong2017metapath2vec}: This approach learns embedding vectors for meta path in heterogeneous networks.\\\vspace{-0.12in}
    \item \textbf{HERec} \cite{shi2018heterogeneous}: This method models recommendation and its auxiliary information as a heterogeneous information network.\\\vspace{-0.12in}
    \item \textbf{HetGNN} \cite{zhang2019heterogeneous}: It combines a two-module architecture and random walk sampling process for heterogeneous graph learning.\\\vspace{-0.12in}
    \item \textbf{HGT}~\cite{hu2020heterogeneous}: This is a graph transformer architecture for encoding diverse relationships in heterogeneous graphs.\\\vspace{-0.12in}
    \item \textbf{HAN} \cite{wang2019heterogeneous}: This method uses a hierarchical attention mechanism to aggregate information from nodes of multiple meta-paths.\\\vspace{-0.12in}
    \item\textbf{DMGI} \cite{park2020unsupervised}: This heterogeneous graph learning method maximizes the mutual information between local and global views. 
    \\\vspace{-0.12in}
    \item \textbf{HeCo} \cite{wang2021self}: It proposes a cross-view contrastive learning approach between two heterogeneous information networks.
\end{itemize}

\noindent\textbf{Homogeneous Graph Convolutional Networks}:\vspace{-0.05in}
\begin{itemize}[leftmargin=*]
    \item \textbf{GraphSage} \cite{hamilton2017inductive}: This model accomplishes graph node aggregation using a node-centric neighbor sampling approach.\\\vspace{-0.12in}
    \item \textbf{GCN} \cite{kipf2016semi}: It aggregates the neighboring nodes embedded after linear projection and finally performs the average computation.\\\vspace{-0.12in}
    \item \textbf{GAE} \cite{kipf2016variational}: This is a variational auto-encoding approach for unsupervised graph representation learning.\\\vspace{-0.12in}
    \item \textbf{GAT} \cite{velivckovic2017graph}: This is a graph attention network that differentiates weights of different neighbors for the central nodes.\\\vspace{-0.12in}
    \item \textbf{LightGCN} \cite{he2020lightgcn}: It employs a lightweight graph convolutional network architecture for the effective aggregation.\\\vspace{-0.12in}
    \item \textbf{DGI} \cite{velivckovic2018deep}: This method learns node embeddings by maximizing the mutual information between local and global graph views.\\\vspace{-0.12in}
    \item \textbf{GraphMAE} \cite{velivckovic2018deep}: It applies the masked auto-encoding training objective to learn node embeddings for downstream tasks.\\\vspace{-0.12in}

    \item \textbf{PinSAGE}~\cite{ying2018graph}: This approach facilitate efficient graph neural networks for web-scale recommendataion.\\\vspace{-0.12in}
    \item \textbf{NGCF}~\cite{wang2019neural}: This is one of the early work that applies graph neural networks to modeling recommendation graphs.
\end{itemize}

\noindent\textbf{Non-Graph Neural Networks}: \vspace{-0.05in}
\begin{itemize}[leftmargin=*]
    \item \textbf{BPR}~\cite{rendle2009bpr}: This baseline enhances matrix factorization by utilizing the Bayesian personalized ranking loss.\\\vspace{-0.12in}
    \item \textbf{DNN}: This is a deep neural network for node classification.
\end{itemize}

\subsubsection{\bf Evaluation Protocols}
The evaluation protocols for our \model\ and the baselines in our experiments are as follows:
\begin{itemize}[leftmargin=*]
    \item For link prediction, we used leave-one-out strategy by generating the test set from users' last interacted items under the target behavior. The evaluation metrics used were Recall@20 and NDCG@20, widely adopted in ranking-based evaluations.\\\vspace{-0.12in} 
    \item For node classification using public datasets DBLP and Aminer, we use 20, 40 and 60 labeled nodes for each class as the training set and 1000 nodes as the validation set and test set, respectively. In our experiments, we used the widely used metrics MicroF1, Macro-F1 and AUC for evaluation. \\\vspace{-0.12in} 
    \item For the node classification task on the Industry dataset, it adopts a more realistic data split scheme, in which the training set and the test set are made of users and interactions of different time periods. Both the training set and the test set contain 1 million user nodes, respectively, and the evaluation metric is AUC with application to the prediction of user retention.
\end{itemize}

\subsubsection{\bf Hyperparameter Settings}
The experiments were conducted on a device with an NVIDIA TITAN RTX GPU and an Intel Xeon W-2133 CPU. The hyperparameter settings are as follows:
\begin{itemize}[leftmargin=*]
    \item \textbf{Link Prediction Task}: We adjusted the learning rate among $\{1e-2, 1e-3, 1e-4\}$. For graph models, the depth of graph propagation layers was tuned within $\{1, 2, 3, 4\}$. We use $l_2$ regularization when training, with the strength coefficient being tuned across $\{1e-1, 3e-2, 1e-2, 1e-3\}$. For diffusion-related parameters, we varied the number of noise steps from 0 to 250, and the noise scale was adjusted among $\{1e-3, 1e-4, 1e-5\}$. The dimensionality of embeddings is tuned from the range $\{8, 16, 32, 64\}$, and and the batch size is selected between 512 and 4096.\\\vspace{-0.12in}
    \item \textbf{Node Classification Task}: For the public datasets DBLP and AMiner, we follow the settings of HeCo. We set the hidden embedding size to 64 and searched for the learning rate from 1e-4 to 5e-3. We varied the number of noise steps from 0 to 250, and the noise scale was tuned among $\{1e-4, 5e-5, 1e-5, 5e-6, 1e-6\}$.
\end{itemize}

\subsection{Overall Performance Comparison (RQ1)}

\begin{table*}[t]
    \centering
    \small
    \caption{Overall performance comparison on the link prediction and node classification tasks.}
    \label{tab:topn}
    \vspace{-0.15in}
    \setlength{\tabcolsep}{0.5mm}
    \resizebox{1\textwidth}{!}{%
    \begin{tabular}{c|c|c|c|c|c|c|c|c|c|c|c|c|c|c|c|c|c|c|c|c}
    \hline
    \multicolumn{21}{c}{Link Prediction Performance on Public Datasets}\\
    \hline
    \multirow{2}{*}{Data}                &  \multicolumn{2}{c|}{BPR}    & \multicolumn{2}{c|}{Pinsage} & \multicolumn{2}{c|}{NGCF}   & \multicolumn{2}{c|}{NMTR}   & \multicolumn{2}{c|}{MBGCN}  & \multicolumn{2}{c|}{HGT}    & \multicolumn{2}{c|}{MATN}   & \multicolumn{2}{c|}{\model}    & \multicolumn{2}{c|}{Improv}  & \multicolumn{2}{c}{p-val} \\ 
    \cline{2-21}
    & R@20 & N@20 & R@20 & N@20& R@20 & N@20& R@20 & N@20& R@20 & N@20& R@20 & N@20& R@20 & N@20& R@20 & N@20& R@20 & N@20& R@20 & N@20\\
    \hline
    {Tmall} & 0.0248& 0.0131 & 0.0368 & 0.0156 & 0.0399 & 0.0169& 0.0441 & 0.0192& 0.0419 & 0.0179& 0.0431 & 0.0192& 0.0463 & 0.0197& 0.0589 & 0.0274& 27.21\% & 39.09\%&   9.8-10&    1.8-11    \\ 
    \hline
    {Rocket} & 0.0308 & 0.0237& 0.0423 & 0.0248& 0.0405 & 0.0257& 0.0460& 0.0265 & 0.0492 & 0.0258& 0.0413 & 0.0250& 0.0524 & 0.0302& 0.0620 & 0.0367& 18.32\% & 21.52\%& 1.65-9 &    3.5-9\\ 
    \hline
    {IJCAI} & 0.0051 & 0.0037& 0.0101  & 0.0041& 0.0091 & 0.0035& 0.0108 & 0.0048& 0.0112 & 0.0045& 0.0126 & 0.0051& 0.0136 & 0.0054& 0.0171 & 0.0063& 25.74\%& 16.67\% &    3.1-11&   2.0-4   \\ 
    \hline
    \hline
    \multicolumn{21}{c}{Node Classification Performance on Industrial Dataset, in terms of AUC}\\
    \hline
    \multicolumn{1}{c|}{Data} &  \multicolumn{2}{c|}{DNN} & \multicolumn{2}{c|}{GraphSage} & \multicolumn{2}{c|}{GCN} & \multicolumn{2}{c|}{GAT} & \multicolumn{2}{c|}{LightGCN} & \multicolumn{2}{c|}{HAN} & \multicolumn{2}{c|}{HGT} & \multicolumn{2}{c|}{HeCo} & \multicolumn{2}{c|}{HGMAE} & \multicolumn{2}{c}{\model} \\ 
    \hline
    \multicolumn{1}{c|}{Ind.} & \multicolumn{2}{c|}{0.7778} & \multicolumn{2}{c|}{0.7836} & \multicolumn{2}{c|}{0.7912} & \multicolumn{2}{c|}{0.7871} & \multicolumn{2}{c|}{0.7904} & \multicolumn{2}{c|}{0.7909} & \multicolumn{2}{c|}{0.7982} &\multicolumn{2}{c|}{0.7915}&\multicolumn{2}{c|}{0.7831}& \multicolumn{2}{c}{0.8025} \\ \hline
    \hline
    \multicolumn{21}{c}{Node Classification performance on public datasets}\\
    \end{tabular}%
    }

    {
    \setlength{\tabcolsep}{1.2mm}
    \begin{tabular}{c|c|c|c|c|c|c|c|c|c|c|c|c|c}
        \hline
        Dataset & \multicolumn{2}{c|}{Metric} & GraphSage & GAE & Mp2vec & HERec & HetGNN & HAN & DGI & DMGI & HeCo & GraphMAE & \model \\ \hline
        \multirow{9}{*}{DBLP} & \multirow{3}{*}{Micro-F1} & 20 & 71.44±8.7 & 91.55±0.1 & 89.67±0.1 & 90.24±0.4 & 90.11±1.0 & 90.16±0.9 & 88.72±2.6 & 90.78±0.3 & 91.97±0.2 & 89.31±0.7 & 93.30±0.4 \\
         &  & 40 & 73.61±8.6 & 90.00±0.3 & 89.14±0.2 & 90.15±0.4 & 89.03±0.7 & 89.47±0.9 & 89.22±0.5 & 89.92±0.4 & 90.76±0.3 & 87.80±0.5 & 93.05±0.3 \\
         &  & 60 & 74.05±8.3 & 90.95±0.2 & 89.14±0.2 & 91.01±0.3 & 90.43±0.6 & 90.34±0.8 & 90.35±0.8 & 90.66±0.5 & 91.59±0.2 & 89.82±0.4 & 93.81±0.3 \\ \cline{2-14} 
         & \multirow{3}{*}{Macro-F1} & 20 & 71.97±8.4 & 90.90±0.1 & 88.98±0.2 & 89.57±0.4 & 89.51±1.1 & 89.31±0.9 & 87.93±2.4 & 89.94±0.4 & 91.28±0.2 & 87.94±0.7 & 93.03±0.4 \\
         &  & 40 & 73.69±8.4 & 89.60±0.3 & 88.68±0.2 & 89.73±0.4 & 88.61±0.8 & 88.87±1.0 & 88.62±0.6 & 89.25±0.4 & 90.34±0.3 & 86.85±0.7 & 92.81±0.3 \\
         &  & 60 & 73.86±8.1 & 90.08±0.2 & 90.25±0.1 & 90.18±0.3 & 89.56±0.5 & 89.20±0.8 & 89.19±0.9 & 89.46±0.6 & 90.64±0.3 & 88.07±0.6 & 93.16±0.3 \\ \cline{2-14} 
         & \multirow{3}{*}{Auc} & 20 & 90.59±4.3 & 98.15±0.1 & 97.69±0.0 & 98.21±0.2 & 97.96±0.4 & 98.07±0.6 & 96.99±1.4 & 97.75±0.3 & 98.32±0.1 & 92.23±3.0 & 99.20±0.1 \\
         &  & 40 & 91.42±4.0 & 97.85±0.1 & 97.08±0.0 & 97.93±0.1 & 97.70±0.3 & 97.48±0.6 & 97.12±0.4 & 97.23±0.2 & 98.06±0.1 & 91.76±2.5 & 98.84±0.1 \\
         &  & 60 & 91.73±3.8 & 98.37±0.1 & 98.00±0.0 & 98.49±0.1 & 97.97±0.2 & 97.96±0.5 & 97.76±0.5 & 97.72±0.4 & 98.59±0.1 & 91.63±2.5 & 99.21±0.1 \\ \hline
        \multirow{9}{*}{AMiner} & \multirow{3}{*}{Micro-F1} & 20 & 49.68±3.1 & 65.78±2.9 & 60.82±0.4 & 63.64±1.1 & 61.49±2.5 & 68.86±4.6 & 62.39±3.9 & 63.93±3.3 & 78.81±1.3 & 68.21±0.3 & 80.55±0.6 \\
         &  & 40 & 52.10±2.2 & 71.34±1.8 & 69.66±0.6 & 71.57±0.7 & 68.47±2.2 & 76.89±1.6 & 63.87±2.9 & 63.60±2.5 & 80.53±0.7 & 74.23±0.2 & 83.29±1.3 \\
         &  & 60 & 51.36±2.2 & 67.70±1.9 & 63.92±0.5 & 69.76±0.8 & 65.61±2.2 & 74.73±1.4 & 63.10±3.0 & 62.51±2.6 & 82.46±1.4 & 72.28±0.2 & 82.10±1.0 \\ \cline{2-14} 
         & \multirow{3}{*}{Macro-F1} & 20 & 42.46±2.5 & 60.22±2.0 & 54.78±0.5 & 58.32±1.1 & 50.06±0.9 & 56.07±3.2 & 51.61±3.2 & 59.50±2.1 & 71.38±1.1 & 62.64±0.2 & 71.98±1.0 \\
         &  & 40 & 45.77±1.5 & 65.66±1.5 & 64.77±0.5 & 64.50±0.7 & 58.97±0.9 & 63.85±1.5 & 54.72±2.6 & 61.92±2.1 & 73.75±0.5 & 68.17±0.2 & 75.57±1.2 \\
         &  & 60 & 44.91±2.0 & 63.74±1.6 & 60.65±0.3 & 65.53±0.7 & 57.34±1.4 & 62.02±1.2 & 55.45±2.4 & 61.15±2.5 & 75.80±1.8 & 68.21±0.2 & 74.57±0.7 \\ \cline{2-14} 
         & \multirow{3}{*}{Auc} & 20 & 70.86±2.5 & 85.39±1.0 & 81.22±0.3 & 83.35±0.5 & 77.96±1.4 & 78.92±2.3 & 75.89±2.2 & 85.34±0.9 & 90.82±0.6 & 86.29±4.1 & 90.19±0.7 \\
         &  & 40 & 74.44±1.3 & 88.29±1.0 & 88.82±0.2 & 88.70±0.4 & 83.14±1.6 & 80.72±2.1 & 77.86±2.1 & 88.02±1.3 & 92.11±0.6 & 89.98±0.0 & 94.41±0.8 \\
         &  & 60 & 74.16±1.3 & 86.92±0.8 & 85.57±0.2 & 87.74±0.5 & 84.77±0.9 & 80.39±1.5 & 77.21±1.4 & 86.20±1.7 & 92.40±0.7 & 88.32±0.0 & 94.25±0.9 \\ \hline
        \end{tabular}
    }    
    
    \vspace{-0.15in}
\end{table*}

This section examines whether \model\ outperforms existing baselines in different graph tasks. The results are presented in Table~\ref{tab:topn}. From the outcomes, we draw the following conclusions:
\begin{itemize}[leftmargin=*]
    \item \textbf{Superior performance of \model}. 
    Across both link prediction and node classification, \model\ consistently outperforms all baselines, demonstrating superior performance. These results underscore the advanced graph forecasting capabilities of the \model\ framework. Additionally, the performance advantages of \model\ on the industry dataset further validate its effectiveness in real-world applications.
    We attribute this success to the model's ability to eliminate inaccurate and spurious semantics from heterogeneous graph structures through the latent heterogeneous graph diffusion paradigm. This semantic refactoring process enables the model to learn more precise and meaningful heterogeneous relations, allowing \model\ to effectively mine valuable heterogeneous information and enhance specific tasks.\\\vspace{-0.12in}
    \item \textbf{Effectiveness of heterogeneous relationships}. 
    The results indicate that methods based on heterogeneous graph learning~\cite{jin2020multi,hu2020heterogeneous} generally outperform those using traditional homogeneous graph neural networks~\cite{wang2019neural,kipf2016semi,velivckovic2017graph}. This advantage highlights the positive role of auxiliary heterogeneous information in graph forecasting tasks. By referring to more information with heterogeneous semantics, graph models can capture richer structural information to improve the prediction accuracy. This advantage is evident in both link prediction and node classification.
    \\\vspace{-0.12in}
    \item \textbf{Effectiveness of hidden-diffusion-based denoising}. The performance of \model\ surpasses that of state-of-the-art heterogeneous graph learning networks, a superiority attributed to the efficacy of our hidden-space diffusion paradigm in denoising and accurately reconstructing complex heterogeneous graph relationships. The denoised source graph heterogeneous semantics serve as a valuable aid in modeling the target graph relations, enabling stronger optimization for the final prediction task.
\end{itemize}

\subsection{Ablation Study (RQ2)}

\begin{figure}[t]
\centering 
    \subfigure[Tmall-Recall]{
        \includegraphics[width=0.32\columnwidth]{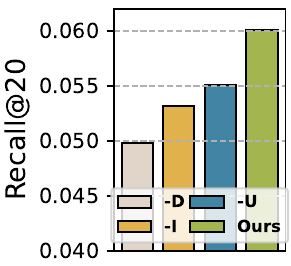}
    } 
    \hspace{-0.1in}
    \subfigure[Rocket-Recall]{
        \includegraphics[width=0.32\columnwidth]{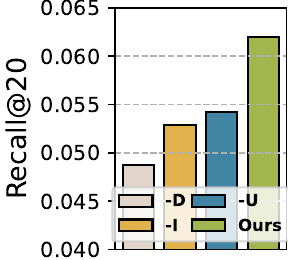}
    } 
    \hspace{-0.1in}
    \subfigure[IJCAI-Recall]{
        \includegraphics[width=0.32\columnwidth]{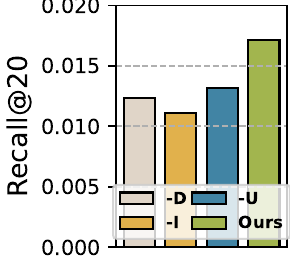}
    }

    \vspace{-0.15in}
    \caption{Ablation study for modules in \model.}
    \label{fig:ablation}
    \vspace{-0.25in}
\end{figure}
To study the impact of \model's sub-modules, we remove or replace essential modules and evaluate the resulting performance. These ablated models include:
\textbf{i) -D:} Omits the holistic diffusion module.
\textbf{ii) -U:} Disregards user-side information of recommendation data.
\textbf{iii) -I:} Similar to -U, eliminates item-side information.
\textbf{iv) -H:} Excludes auxiliary heterogeneous graphs when training.
\textbf{v) DAE:} Replace the diffusion module with a denoising autoencoder.

The evaluation results for link prediction is listed in Figure~\ref{fig:ablation}. We also conduct ablation study for node classification on the Industry data. The results in terms of AUC, is as follows: \textbf{-H}, 0.7840, \textbf{-D}, 0.7901, \textbf{DAE}, 0.7911, \textbf{\model}, 0.8025.
Through meticulous examination, we make the following noteworthy observations:
\begin{itemize}[leftmargin=*]
    \item Removing the diffusion module "-D" leads to performance degradation, highlighting both the adverse effects of noise in auxiliary heterogeneous data and demonstrating the effectiveness of our latent feature-level diffusion model in its denoising function.
    \item Comparing "-U" to \model\ demonstrates the value of heterogeneous user aggregation. Similarly, "-I" and "-H" variants confirm the benefits of heterogeneous information. Notably, "-D" outperforming "-I" suggests that removing item-side diffusion both loses valuable signals and introduces noise.
    \item The "DAE" variant shows only limited performance improvement over completely removing the denoising module (\ie, "-D"). This demonstrates the difficulty of denoising heterogeneous graph data using existing DAE approaches. Consequently, it validates the superiority of the stepwise noise addition and removal process in our latent heterogeneous graph diffusion module.
\end{itemize}
\begin{figure}[t]
    \centering
        
    \subfigure[GCN Layers]{
        \includegraphics[width=0.32\columnwidth]{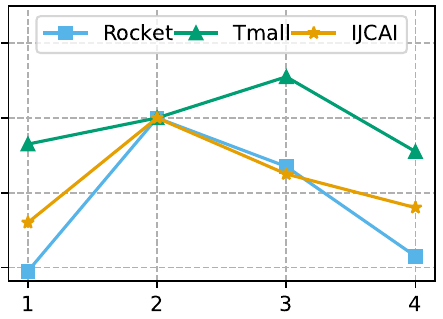}
    }
    \hspace{-0.1in}
    \subfigure[Dimensionality $d$]{
        \includegraphics[width=0.32\columnwidth]{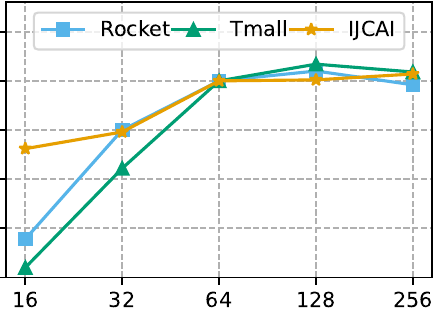}
    }
    \hspace{-0.1in}
    \subfigure[Diffusion Steps $t$]{
        \includegraphics[width=0.32\columnwidth]{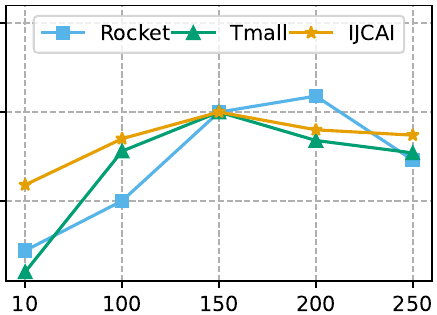}
    }
    \vspace{-0.15in}
    \caption{Hyperparameter study in terms of Recall@20.}
    \label{fig:hyper}
    \vspace{-0.2in}
\end{figure}

\subsection{Impact of Hyperparameters (RQ3)}

This section examines the impact of hyperparameter settings. The results are presented in Figure~\ref{fig:hyper}. We make the following analysis.
\begin{itemize}[leftmargin=*]

    \item \textbf{Graph propagation iterations} $L$.  Tested within {1, 2, 3, 4}. Performance improves up to $L = 3$; beyond this, additional layers may introduce noise and result in over-smoothing issues~\cite{chen2020measuring,xia2020multiplex}.
    
    \item \textbf{Embedding dimensionality} $d$. anges tested were {16, 32, 64, 128, 256}. As $d$ increases, there's a general improvement in performance, peaking before $d = 256$, where performance slightly declines due to potential overfitting.

    \item \textbf{Maximum diffusion steps} $T$. Varied from 10 to 250. Increasing the diffusion steps $T$ typically enhances the performance for \model, but very high values degrade it, likely due to the excessive noise disrupting social information integrity.

    \item \textbf{Noise scale} $S$. Evaluated at scales {1e-3, 1e-4, 1e-5, 1e-6} with respective AUC scores of 0.7966, 0.8025, 0.7988, 0.7974 on the Industry dataset. Optimal noise scales improve model performance by enhancing denoising effectiveness; however, excessively high scales diminish performance by obscuring key graph semantics.

\end{itemize}

\subsection{Efficiency Study (RQ4)}

\begin{figure}
    \centering
    \includegraphics[width=0.7\columnwidth]{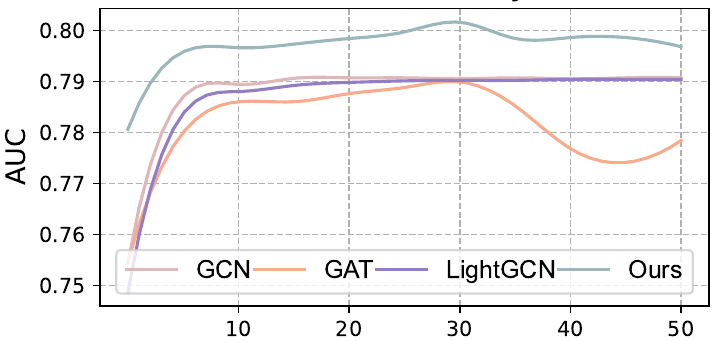}
    \vspace{-0.15in}
    \caption{Test performance v.s. training epochs.}
    \label{fig:efficient}
    \vspace{-0.18in}
\end{figure}

This section assesses the computational efficiency of \model. On industry data, we conducted a convergence analysis shown in Figure~\ref{fig:efficient}. We also compared the per-epoch running time of \model\ and baselines on two link datasets with the following results:
\textbf{Tmall:} \textbf{HGT}, 16.824s, \textbf{MBGCN}, 12.644s, \textbf{\model}, 6.558s. \textbf{Retail\_rocket:} 
\textbf{HGT}, 2.451s, \textbf{MBGCN}, 2.312s, \textbf{\model}, 1.811s. 
The results show  \model\ consistently outperforms baselines in training efficiency, benefiting from its effective hidden diffusion.

\vspace{-0.03in}
\subsection{Performance w.r.t. Data Sparsity (RQ5)}
\begin{figure}
    \centering
    \includegraphics[width=0.49\columnwidth]{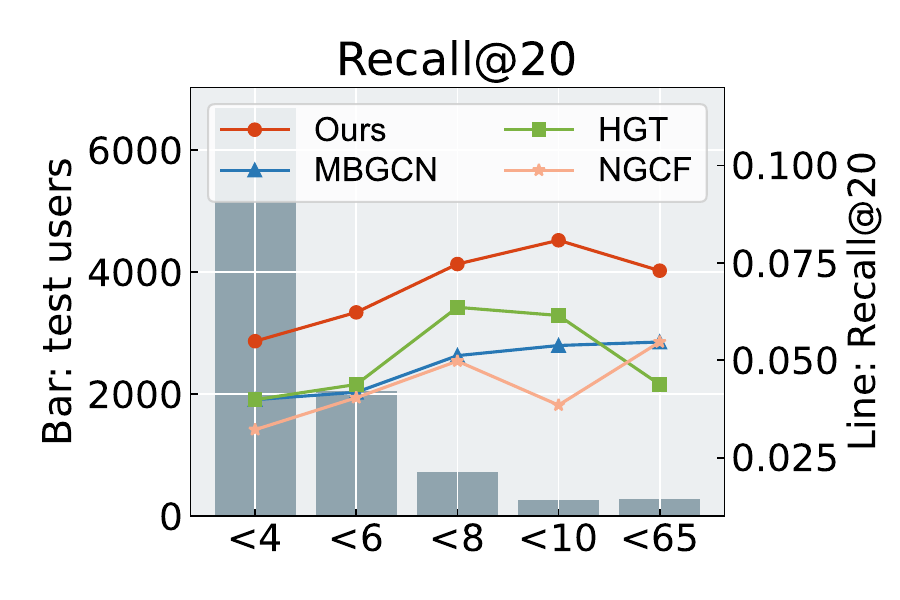}
    \hspace{-0.05in}
    \includegraphics[width=0.49\columnwidth]{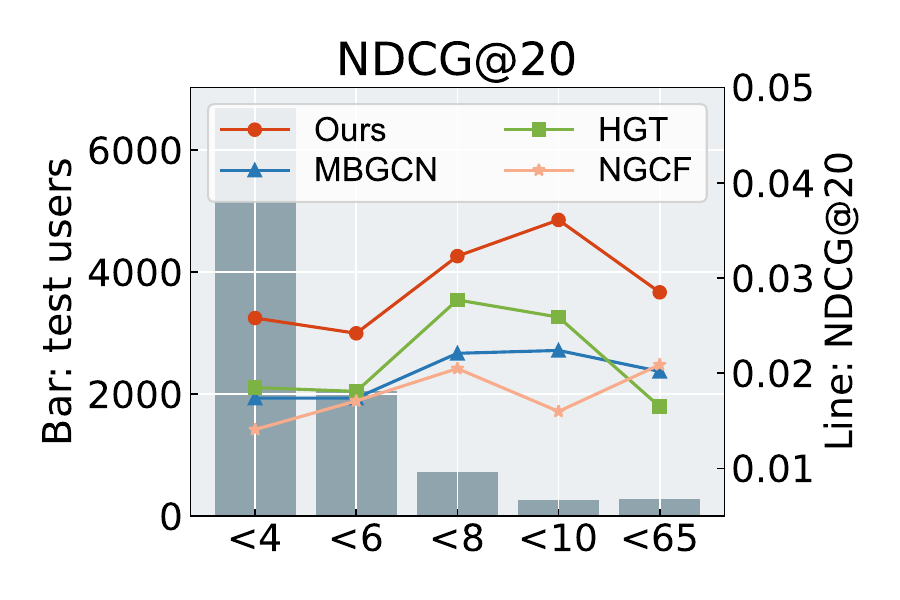}
    \vspace{-0.15in}
    \caption{Performance \textit{w.r.t.} different data sparsity}
    \label{fig:sparse}
    \vspace{-0.2in}
\end{figure}

We further assess how the diffusion model in \model\ address the data sparsity issue in heterogeneous graph learning. For the Tmall dataset, we split nodes into five groups based on interaction counts (\eg, "<8", "<65"), and evaluate the performance on each group. The results are shown in Figure~\ref{fig:sparse}. Our observations include:
\begin{itemize}[leftmargin=*]
    \item \textbf{Trend in Performance}. As edges per node increase, so does performance across all methods, indicating better embeddings from richer edge data. However, performance dips in groups with "<10" and "<65" links, likely due to fewer test samples.
    \item \textbf{Effect of Heterogeneity Learning}. Heterogeneous models (MBGCN, HGT) outperform the homogeneous NGCF,
    showing that multiple context fusion effectively combats data sparsity.
    \item \textbf{Superiority of \model}. \model\ consistently surpasses all baselines, suggesting its diffusion-based approach effectively preserves and utilizes heterogeneous behavioral data, thus offering a robust solution to data sparsity challenges.
\end{itemize}

\subsection{Exploration of Anti-Noise Capacity (RQ6)}
\begin{table}[t]
    \centering
    \small
    \setlength{\tabcolsep}{1.2mm}
    
    \caption{Performance decay caused by different noise ratios.}
    \label{tab:noise_ratio}
    \vspace{-0.15in}
        \begin{tabular}{cc|cc|cc|cc}
            \hline
            \multicolumn{2}{c|}{\multirow{2}{*}{Noise Ratios}} & \multicolumn{2}{c|}{10\%} & \multicolumn{2}{c|}{30\%} & \multicolumn{2}{c}{50\%} \\ \cline{3-8} 
            \multicolumn{2}{c|}{} & Recall & NDCG & Recall & NDCG & Recall & NDCG \\ \hline
            \multicolumn{1}{c|}{\multirow{2}{*}{pv}} & \model & 98.28\% & 96.31\% & 98.80\% & 97.05\% & 97.42\% & 96.68\% \\
            \multicolumn{1}{c|}{} & HGT & 96.98\% & 94.79\% & 97.45\% & 96.35\% & 95.59\% & 90.10\% \\ \hline
            \multicolumn{1}{c|}{\multirow{2}{*}{fav}} & \model & 97.93\% & 95.57\% & 99.14\% & 95.57\% & 98.62\% & 96.31\% \\
            \multicolumn{1}{c|}{} & HGT & 98.14\% & 93.75\% & 97.68\% & 91.67\% & 97.22\% & 89.06\% \\ \hline
            \multicolumn{1}{c|}{\multirow{2}{*}{cart}} & \model & 98.97\% & 98.15\% & 98.28\% & 97.42\% & 96.73\% & 92.62\% \\
            \multicolumn{1}{c|}{} & HGT & 98.61\% & 96.88\% & 96.29\% & 95.31\% & 95.82\% & 92.19\% \\ \hline
        \end{tabular}%
        \vspace{-0.15in}
\end{table}
To study the model robustness against data noise, we evaluate the percentage of performance degradation for each heterogeneous relation under different noise ratios (Table~\ref{tab:noise_ratio}). In our experiment, 0, 10$\%$, 30$\%$ and 50$\%$ of the heterogeneous relations are randomly replaced by noise signals. We tested the performance of \model\ and HGT on the Tmall dataset. The results demonstrate that our model experiences a markedly less pronounced decline in performance in comparison to the baseline model. Such findings substantiate the enhanced denoising capability of our \model\ approach.

\subsection{Case Study (RQ7)}
We utilized the t-SNE technique to visualize the embeddings of different relation types from the Tmall dataset, color-coding each type for distinction. By comparing the behavior embedding of our \model\ to its variant "-D" (which lacks the denoising module), it's evident that our model achieves clearer separation, drawing embeddings with similar semantics closer together. This demonstrates our model's proficiency in capturing and refining heterogeneous data through hidden-space diffusion and denoising, effectively reducing noise and enhancing semantic accuracy. By optimizing the semantic diffusion from source to target, our method efficiently processes and highlights diverse information across graph structures, significantly improving learning outcomes.

\begin{figure}
    \centering
    \includegraphics[width=0.46\columnwidth]{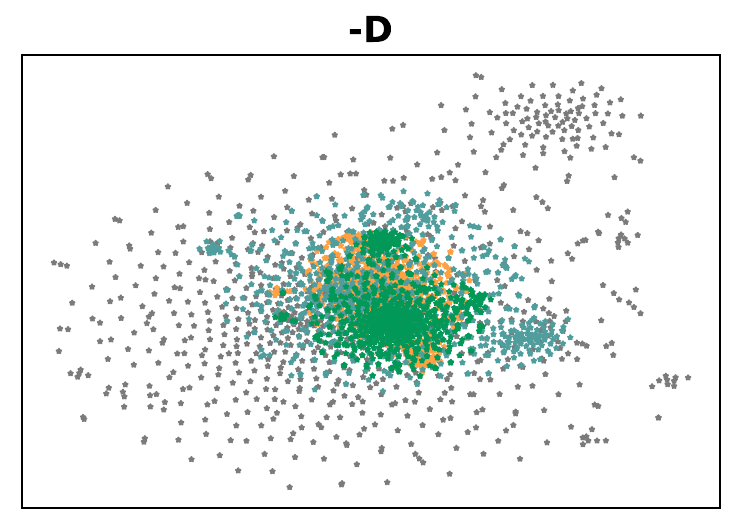}
    \,\,
    \includegraphics[width=0.46\columnwidth]{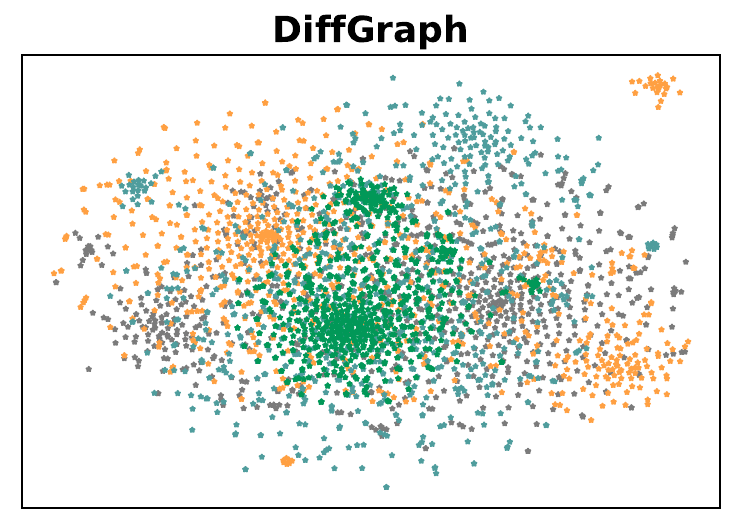}
    \vspace{-0.1in}
    \caption{Embedding visualization on the Tmall dataset.}
    \label{fig:case}
    \vspace{-0.12in}
\end{figure}
\section{Related Work}
\noindent\textbf{Graph Neural Networks}.  
Graph learning has evolved substantially, with GNNs demonstrating significant impact across domains including spammer detection~\cite{li2019spam, wu2020graph, zhang2020gcn}, recommender systems~\cite{yu2021self, fan2019graph}, and machine translation~\cite{luong2015effective, wu2016google}. While traditional GNNs like Graph Convolutional Networks (GCNs)~\cite{gao2018large, zhang2021lorentzian} and Graph Attention Networks (GATs)~\cite{zhang2022graph, liao2019efficient} have advanced representation learning on graphs, they primarily focus on homogeneous or simple bipartite structures. To address the limitations in modeling complex heterogeneous graph relations, we propose a denoised heterogeneous graph neural network based on the diffusion model. \\\vspace{-0.12in}

\noindent\textbf{Heterogeneous Graph Learning}.
Heterogeneous graphs, featuring diverse node and link types, are crucial in various real-world applications. Heterogeneous graph neural networks (HGNNs) develop node embeddings that capture complex semantics for tasks like node classification, edge classification, and link prediction~\cite{zhang2019heterogeneous,tang2024higpt}. Methods like HetGNN~\cite{zhang2019heterogeneous} and MAGNN~\cite{fu2020magnn} incorporate attention mechanisms and meta-path aggregations to refine these embeddings. HGT~\cite{hu2020heterogeneous} leverages a transformer-based architecture for modeling diverse relationships. While these methods enhance the capture of heterogeneous semantics, they often ignore noise and irrelevant information. Our model introduces a diffusion-based denoising mechanism to extract genuine, task-relevant information.\\\vspace{-0.12in}

\noindent\textbf{Generative Models for Graph Learning}.  
Graph generation aims to uncover structural patterns, mitigate anomalies, and simulate new graphs, garnering significant interest. Traditional models~\cite{albert2002statistical, kleinberg2000small, erdds1959random} were limited to generating graphs with specific statistical traits. However, advancements in deep learning have expanded possibilities. Techniques like Generative Adversarial Networks (GANs)~\cite{wang2017irgan} and Variational Autoencoders (VAEs)~\cite{Yu_Zhang_Cao_Xia} play key roles in data generation and graph task learning. Recent efforts, including diffusion models~\cite{austin2021structured, Croitoru_Hondru_Ionescu_Shah_2022, popov2021grad}, focus on generative recommendation with enhanced capabilities. Despite these advancements, the use of these models in graph denoising is limited. This work pioneers the application of generative models to transition from auxiliary heterogeneous graphs to targeted semantic spaces.
\section{Conclusion}
\label{sec:conclusion}
This paper presents \model, a heterogeneous graph diffusion model that advances graph learning through two key innovations: 1) A latent diffusion mechanism that progressively filters noise while preserving task-relevant signals from source data, and 2) An enhanced semantic transition framework that better captures relationships across different interaction types, enabling more nuanced heterogeneous graph modeling. Extensive experiments conducted on both public benchmarks and industrial datasets demonstrate our approach's superiority in terms of prediction accuracy and model robustness. The results consistently validate the effectiveness of our proposed techniques across various scenarios. Future work could explore extending our model to dynamic heterogeneous graphs, where both node attributes and graph structure evolve over time.

\clearpage
\bibliographystyle{abbrv}
\balance
\bibliography{refs}

\clearpage
\section*{Ethical Considerations}
In this section, we examine the ethical implications of the \model\ framework, which is designed to distill essential information from auxiliary relational data for enhancing prediction tasks in heterogeneous graphs. Our discussion addresses potential ethical challenges and considerations that may arise during both the implementation and deployment phases of the \model\ framework. \\\vspace{-0.1in}

\noindent\textbf{Data Privacy and Consent}.
The \model\ framework processes auxiliary relational data to enhance predictions in heterogeneous graphs, which inherently raises important privacy and consent considerations. When handling relational data, particularly those containing personal information, strict compliance with data protection regulations (e.g., GDPR, CCPA) is paramount. The potential for unintended inference or exposure of sensitive information through relational analysis presents a significant privacy concern, especially if explicit user consent is not properly obtained. To address these challenges, implementations must incorporate robust safeguards, including strict access control mechanisms, advanced data anonymization techniques, and transparent data usage policies. These measures are essential for protecting user privacy while maintaining the framework's analytical capabilities. \\\vspace{-0.1in}

\noindent\textbf{Transparency and Explainability}.
The sophisticated nature of the \model\ framework, particularly its mechanisms for integrating and interpreting auxiliary relational data, presents significant challenges in terms of transparency and explainability. In high-stakes environments, stakeholders need clear understanding of how the model arrives at its predictions and subsequent recommendations. The inherent complexity of the model can potentially obscure decision-making processes, making it difficult to identify and rectify logical errors while potentially eroding stakeholder trust. To address these challenges, it is crucial to develop robust explainability methods, such as interpretable feature importance metrics and comprehensive decision rationale documentation. These transparency measures not only foster trust among stakeholders but also ensure compliance with regulatory requirements and maintain accountability throughout the model's deployment lifecycle.


\end{document}